\documentclass[11pt,a4paper]{article}
\usepackage[hyperref]{acl2017}
\usepackage{times}
\usepackage{latexsym}
\usepackage{graphicx}
\usepackage{subcaption}
\usepackage{amssymb}
\usepackage{amsmath}
\usepackage{framed}
\usepackage{fancyvrb}
\usepackage{soul}
\usepackage{url}
\usepackage{color}

\aclfinalcopy % Uncomment this line for the final submission
\title{Hybrid Code Networks: practical and efficient end-to-end dialog control with supervised and reinforcement learning}

\author{Jason D. Williams \\
  Microsoft Research \\
  {\small \tt jason.williams@microsoft.com} \\
  \\\And
  Kavosh Asadi \\
  Brown University \\
  {\small \tt kavosh@brown.edu}  \\\And
  Geoffrey Zweig\thanks{\hspace{1mm} Currently at JPMorgan Chase} \\
  Microsoft Research \\
  {\small \tt g2zweig@gmail.com} \\}

\date{}

\begin{document}
\maketitle
\begin{abstract}
End-to-end learning of recurrent neural networks (RNNs) is an attractive solution for dialog systems; however, current techniques are data-intensive and require thousands of dialogs to learn simple behaviors.  We introduce Hybrid Code Networks (HCNs), which combine an RNN with \emph{domain-specific knowledge encoded as software} and \emph{system action templates}.  Compared to existing end-to-end approaches, HCNs considerably reduce the amount of training data required, while retaining the key benefit of inferring a latent representation of dialog state.  In addition, HCNs can be optimized with supervised learning, reinforcement learning, or a mixture of both.  HCNs attain state-of-the-art performance on the bAbI dialog dataset \citep{dialogbabiarxiv}, and outperform two commercially deployed customer-facing dialog systems.
\end{abstract}

%The RNN maps from raw dialog history directly to a distribution over system action templates, and the software tracks entities, provides API access, and expresses constraints over actions.  

\section{Introduction}
\label{sec:intro}

Task-oriented dialog systems help a user to accomplish some goal using natural language, such as making a restaurant reservation, getting technical support, or placing a phonecall.  Historically, these dialog systems have been built as a pipeline, with modules for language understanding, state tracking, action selection, and language generation.  However, dependencies between modules introduce considerable complexity -- for example, it is often unclear how to define the dialog state and what history to maintain, yet action selection relies exclusively on the state for input.  Moreover, training each module requires specialized labels.

Recently, end-to-end approaches have trained recurrent neural networks (RNNs) directly on text transcripts of dialogs.  A key benefit is that the RNN infers a latent representation of state, obviating the need for state labels.  However, end-to-end methods lack a general mechanism for injecting domain knowledge and constraints.  For example, simple operations like sorting a list of database results or updating a dictionary of entities can expressed in a few lines of software, yet may take thousands of dialogs to learn.  Moreover, in some practical settings, programmed constraints are essential -- for example, a banking dialog system would require that a user is logged in before they can retrieve account information.  

%This paper presents a model for end-to-end learning -- \emph{Hybrid Code Networks} (HCNs) -- which learns end-to-end but also allows a developer to express domain knowledge in software.  Because the RNN infers its own representation of state, little or no hand-crafting of state is required, apart from tracking entity values.  
This paper presents a model for end-to-end learning, called \emph{Hybrid Code Networks} (HCNs) which addresses these problems.  In addition to learning an RNN, HCNs also allow a developer to express domain knowledge via software and action templates.  Experiments show that, compared to existing recurrent end-to-end techniques, HCNs achieve the same performance with considerably less training data, while retaining the key benefit of end-to-end trainability.  Moreover, the neural network can be trained with supervised learning or reinforcement learning, by changing the gradient update applied.
  
This paper is organized as follows.  Section \ref{sec:model} describes the model, and Section \ref{sec:related} compares the model to related work.  Section \ref{sec:sl1} applies HCNs to the bAbI dialog dataset \cite{dialogbabiarxiv}.  Section \ref{sec:sl2} then applies the method to real customer support domains at our company.  Section \ref{sec:rl} illustrates how HCNs can be optimized with reinforcement learning, and Section \ref{sec:concl} concludes.

\section{Model description}
\label{sec:model}

\begin{figure*}[t]
\begin{center}
\includegraphics[trim = 45mm 60mm 60mm 30mm, scale=0.59]{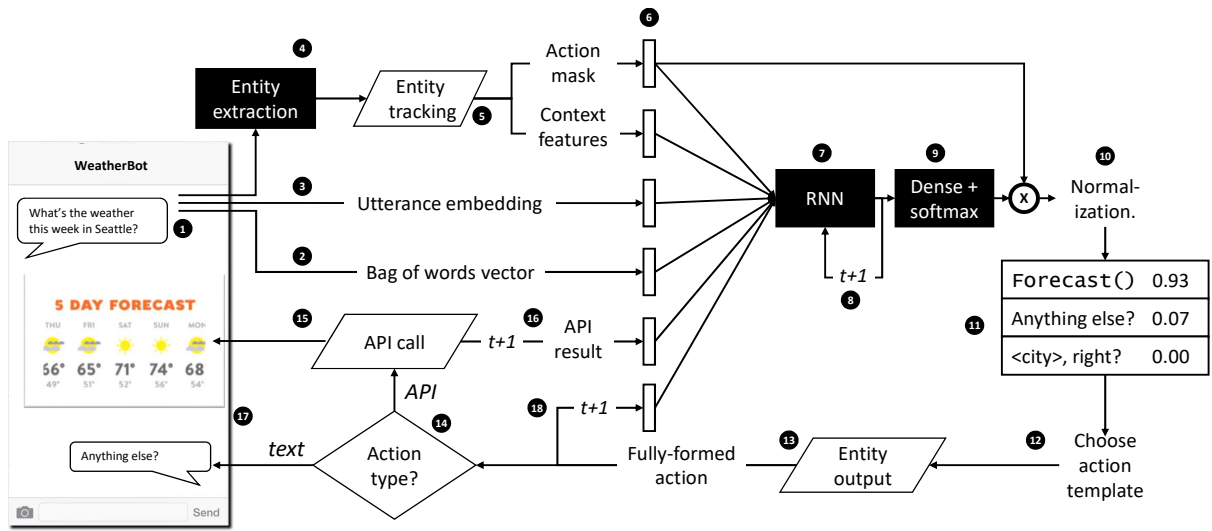}
\caption{\label{fig:diagram}Operational loop.  Trapezoids refer to programmatic code provided by the software developer, and shaded boxes are trainable components.  Vertical bars under ``6'' represent concatenated vectors which form the input to the RNN.}
\end{center}
\end{figure*}

At a high level, the four components of a Hybrid Code Network are a recurrent neural network; domain-specific software; domain-specific action templates; and a conventional entity extraction module for identifying entity mentions in text.  Both the RNN and the developer code maintain state.  Each action template can be a textual communicative action or an API call.  The HCN model is summarized in Figure \ref{fig:diagram}.

The cycle begins when the user provides an utterance, as text (step 1).  The utterance is featurized in several ways.  First, a bag of words vector is formed (step 2).  Second, an utterance embedding is formed, using a pre-built utterance embedding model (step 3).  Third, an entity extraction module identifies entity mentions (step 4) -- for example, identifying ``Jennifer Jones'' as a \verb|<name>| entity.  The text and entity mentions are then passed to ``Entity tracking'' code provided by the developer (step 5), which grounds and maintains entities -- for example, mapping the text ``Jennifer Jones'' to a specific row in a database.  This code can optionally return an ``action mask'', indicating actions which are permitted at the current timestep, as a bit vector.  For example, if a target phone number has not yet been identified, the API action to place a phone call may be masked.  It can also optionally return ``context features'' which are features the developer thinks will be useful for distinguishing among actions, such as which entities are currently present and which are absent.

The feature components from steps 1-5 are concatenated to form a feature vector (step 6).  This vector is passed to an RNN, such as a long short-term memory (LSTM) \citep{lstm} or gated recurrent unit (GRU) \citep{gru}.  The RNN computes a hidden state (vector), which is retained for the next timestep (step 8), and passed to a dense layer with a softmax activation, with output dimension equal to the number of distinct system action templates (step 9).\footnote{Implementation details for the RNN such as size, loss, etc. are given with each experiment in Sections \ref{sec:sl1}-\ref{sec:rl}.}  Thus the output of step 9 is a distribution over action templates.  Next, the action mask is applied as an element-wise multiplication, and the result is normalized back to a probability distribution (step 10) -- this forces non-permitted actions to take on probability zero.  From the resulting distribution (step 11), an action is selected (step 12).  When RL is active, exploration is required, so in this case an action is \emph{sampled} from the distribution; when RL is not active, the best action should be chosen, and so the action with the \emph{highest probability} is always selected.

The selected action is next passed to ``Entity output'' developer code that can substitute in entities (step 13) and produce a fully-formed action -- for example, mapping the template ``\verb|<city>|, right?'' to ``Seattle, right?''.  In step 14, control branches depending on the type of the action: if it is an API action, the corresponding API call in the developer code is invoked (step 15) -- for example, to render rich content to the user.  APIs can act as sensors and return features relevant to the dialog, so these can be added to the feature vector in the next timestep (step 16).  If the action is text, it is rendered to the user (step 17), and cycle then repeats.  The action taken is provided as a feature to the RNN in the next timestep (step 18).

\section{Related work}
\label{sec:related}

Broadly there are two lines of work applying machine learning to dialog control.  The first decomposes a dialog system into a pipeline, typically including language understanding, dialog state tracking, action selection policy, and language generation \citep{levin2000stochastic,singh2002optimizing,williams2007csl,williams2008icslp1,4960703,lee2009example,griol2008,young2013ieee,li2014slt}.  
%By extracting features from the dialog state, action selection can be framed as a compact decision problem, admitting techniques such as Markov decision processes (MDP) \citep{levin2000stochastic,singh2002optimizing}, partially observable Markov decision processes (POMDPs) \citep{williams2007csl,young2013ieee}, or supervised learning \citep{4960703,lee2009example,griol2008,li2014slt}.  
Specifically related to HCNs, past work has implemented the policy as feed-forward neural networks \citep{DBLP:journals/corr/WenGMRSUVY16}, trained with supervised learning followed by reinforcement learning \citep{su:2016:nnpolicy}.  In these works, the policy has not been \emph{recurrent} -- i.e., the policy depends on the state tracker to summarize observable dialog history into state features, which requires design and specialized labeling.  By contrast, HCNs use an RNN which automatically infers a representation of state.  For learning efficiency, HCNs use an external light-weight process for tracking entity values, but the policy is not strictly dependent on it: as an illustration, in Section \ref{sec:sl2} below, we demonstrate an HCN-based dialog system which has no external state tracker.  If there is context which is not apparent in the text in the dialog, such as database status, this can be encoded as a context feature to the RNN.
%In these works, action selection is entirely dependent on the state tracker to summarize dialog history, such as which slots have been requested, confirmed, denied, etc.  By contrast, HCNs also employ a state tracker, but its task is much simpler: it need store only entity values.  Indeed, it is possible to build HCN-based dialog systems with \emph{no} explicit state tracker: in Section \ref{sec:sl2} below, we demonstrate an HCN-based dialog system without an entity state tracker.  Some dialog context is not apparent in the text in the dialog, such as how many items in the database match; HCNs provide an affordance for context features so this information can be expressed.

The second, more recent line of work applies recurrent neural networks (RNNs) to learn ``end-to-end'' models, which map from an observable dialog history directly to a sequence of output words \citep{sordoni2015,shang2015,vinyals2015,yao2015,Serban2016AAAI,li2016naacl,li2016emnlp,DBLP:journals/corr/LuanJO16,DBLP:journals/corr/XuLWSW16,li2016acl,DBLP:journals/corr/MeiBW16,lowe2017,serban2017}.  These systems can be applied to task-oriented domains by adding special ``API call'' actions, enumerating database output as a sequence of tokens \citep{dialogbabiarxiv}, then learning an RNN using Memory Networks \citep{memnetworksnips}, gated memory networks \citep{GMemNNarxiv}, query reduction networks \citep{QRNarxiv}, and copy-augmented networks \citep{copyaugmented}.  In each of these architectures, the RNN learns to manipulate entity values, for example by saving them in a memory.  Output is produced by generating a sequence of tokens (or ranking all possible surface forms), which can also draw from this memory.  HCNs also use an RNN to accumulate dialog state and choose actions.  However, HCNs differ in that they use developer-provided action templates, which can contain entity references, such as ``\verb|<city>|, right?''.  This design reduce learning complexity, and also enable the software to limit which actions are available via an action mask, at the expense of developer effort.  To further reduce learning complexity in a practical system, entities are tracked separately, outside the the RNN, which also allows them to be substituted into action templates.  Also, past end-to-end recurrent models have been trained using supervised learning, whereas we show how HCNs can also be trained with reinforcement learning. 

%Apart from machine learning approaches, there are numerous examples of rule-based dialog managers, including examples in both research \citep{larsson2000information} and industry \citep{VoiceXML}.  By contrast, HCN can be viewed as a partial program \citep{andre2002}, where the rule-based dialog manager periodically nominates a set of actions for a machine-learning algorithm to select from.  
% In contrast, in HCNs developers inject domain knowledge, in the form of domain-specific software and action templates.

%Finally, there has been at least one past attempt to marry rule-based approaches with machine learning \citep{williams2008icslp1}; in contrast to HCNs, that work required a full state tracker, and was limited to reinforcement learning. 

\section{Supervised learning evaluation I}
\label{sec:sl1}

In this section we compare HCNs to existing approaches on the public ``bAbI dialog'' dataset \citep{dialogbabiarxiv}. This dataset includes two end-to-end dialog learning tasks, in the restaurant domain, called task5 and task6.\footnote{Tasks 1-4 are sub-tasks of Task5.}  Task5 consists of synthetic, simulated dialog data, with highly regular user behavior and constrained vocabulary.  Dialogs include a database access action which retrieves relevant restaurants from a database, with results included in the dialog transcript.  We test on the ``OOV'' variant of Task5, which includes entity values not observed in the training set.  Task6 draws on human-computer dialog data from the second dialog state tracking challenge (DSTC2), where usability subjects (crowd-workers) interacted with several variants of a spoken dialog system \citep{DSTC2Summary}.  Since the database from DSTC2 was not provided, database calls have been inferred from the data and inserted into the dialog transcript.  Example dialogs are provided in the Appendix Sections \ref{sec:task5_ed} and \ref{sec:task6_ed}.

To apply HCNs, we wrote simple domain-specific software, as follows.  First, for entity extraction (step 4 in Figure \ref{fig:diagram}), we used a simple string match, with a pre-defined list of entity names -- i.e., the list of restaurants available in the database.  
%Following past work, we match entity names not included in the training set, since the database must by construction include the names and properties of the restaurants it contains.  
Second, in the context update (step 5), we wrote simple logic for tracking entities: when an entity is recognized in the user input, it is retained by the software, over-writing any previously stored value.  For example, if the \verb|price| ``cheap'' is recognized in the first turn, it is retained as \verb|price=cheap|.  If ``expensive'' is then recognized in the third turn, it over-writes ``cheap'' so the code now holds \verb|price=expensive|.
% created simple rules for tracking entities, where entities recognized in the input overwrite the existing entries.  
Third, system actions were templatized: for example, system actions of the form ``prezzo is a nice restaurant in the west of town in the moderate price range'' all map to the template ``\verb|<name>| is a nice restaurant in the \verb|<location>| of town in the \verb|<price>| price range''.  This results in 16 templates for Task5 and 58 for Task6.\footnote{A handful of actions in Task6 seemed spurious; for these, we replaced them with a special ``UNK'' action in the training set, and masked this action at test time.}
%The HCN RNN outputs a distribution over the templates, and entities are substituted into the templates from the entity dictionary (step 10).  
Fourth, when database results are received into the entity state, they are sorted by rating.  Finally, an action mask was created which encoded common-sense dependencies.  These are implemented as simple if-then rules based on the presence of entity values: for example, only allow an API call if pre-conditions are met; only offer a restaurant if database results have already been received; do not ask for an entity if it is already known; etc.

For Task6, we noticed that the system can say that no restaurants match the current query \emph{without} consulting the database (for an example dialog, see Section \ref{sec:task6_ed} in the Appendix).  In a practical system this information would be retrieved from the database and not encoded in the RNN.  So, we mined the training data and built a table of search queries known to yield no results.  We also added context features that indicated the state of the database -- for example, whether there were any restaurants matching the current query.  The complete set of context features is given in Appendix Section \ref{sec:context_features}.  Altogether this code consisted of about 250 lines of Python.

We then trained an HCN on the training set, employing the domain-specific software described above.  We selected an LSTM for the recurrent layer \citep{lstm}, with the AdaDelta optimizer \citep{DBLP:journals/corr/abs-1212-5701}.  We used the development set to tune the number of hidden units (128), and the number of epochs (12).  Utterance embeddings were formed by averaging word embeddings, using a publicly available 300-dimensional word embedding model trained using word2vec on web data \citep{mikolov2013distributed}.\footnote{Google News 100B model from \url{https://github.com/3Top/word2vec-api}}  The word embeddings were static and not updated during LSTM training.  In training, each dialog formed one minibatch, and updates were done on full rollouts (i.e., non-truncated back propagation through time).  The training loss was categorical cross-entropy.  Further low-level implementation details are in the Appendix Section \ref{sec:impl_details}.

We ran experiments with four variants of our model: with and without the utterance embeddings, and with and without the action mask (Figure \ref{fig:diagram}, steps 3 and 6 respectively).  

Following past work, we report average turn accuracy -- i.e., for each turn in each dialog, present the (true) history of user and system actions to the network and obtain the network's prediction as a string of characters.  The turn is correct if the string matches the reference exactly, and incorrect if not.  We also report dialog accuracy, which indicates if all turns in a dialog are correct.

We compare to four past end-to-end approaches \citep{dialogbabiarxiv,GMemNNarxiv,copyaugmented,QRNarxiv}.  We emphasize that past approaches have applied purely sequence-to-sequence models, or (as a baseline) purely programmed rules \citep{dialogbabiarxiv}.  By contrast, Hybrid Code Networks are a hybrid of hand-coded rules and learned models.

\begin{table*}[t]
\begin{center}
\begin{tabular}{c|cc|cc}
\hline 
 & \multicolumn{2}{|c|}{\bf Task5-OOV} & \multicolumn{2}{|c}{\bf Task6} \\ 
\bf Model & Turn Acc. & Dialog Acc. & Turn Acc. & Dialog Acc. \\
\hline
Rules                   & \bf 100\%  & \bf 100\%  & 33.3\% & 0.0\% \\
\hline
\citet{dialogbabiarxiv} &     77.7\% &     0.0\%  &     41.1\% &     0.0\% \\
\citet{GMemNNarxiv}     &     79.4\% &     0.0\%  &     48.7\% &     1.4\% \\
\citet{copyaugmented}   &     ---    &     ---    &     48.0\% &     1.5\% \\
\citet{QRNarxiv}        &     96.0\% &     ---    &     51.1\% &     ---   \\
\hline
HCN                   &  \bf 100\% &  \bf 100\% &     54.0\% &     1.2\% \\
HCN+embed             &  \bf 100\% &  \bf 100\% & \bf 55.6\% &     1.3\% \\
HCN+mask              &  \bf 100\% &  \bf 100\% &     53.1\% & \bf 1.9\% \\
HCN+embed+mask        &  \bf 100\% &  \bf 100\% &     52.7\% &     1.5\% \\
\hline
\end{tabular}
\end{center}
\caption{\label{tab:sl1} Results on bAbI dialog Task5-OOV and Task6 \citep{dialogbabiarxiv}.  Results for ``Rules'' taken from \citet{dialogbabiarxiv}.  Note that, unlike cited past work, HCNs make use of domain-specific procedural knowledge.}
\end{table*}

\begin{figure*}[t]
    \centering
    \begin{subfigure}[t]{0.5\textwidth}
        \centering
        \includegraphics[trim = 20mm 20mm 20mm 20mm, scale=0.33]{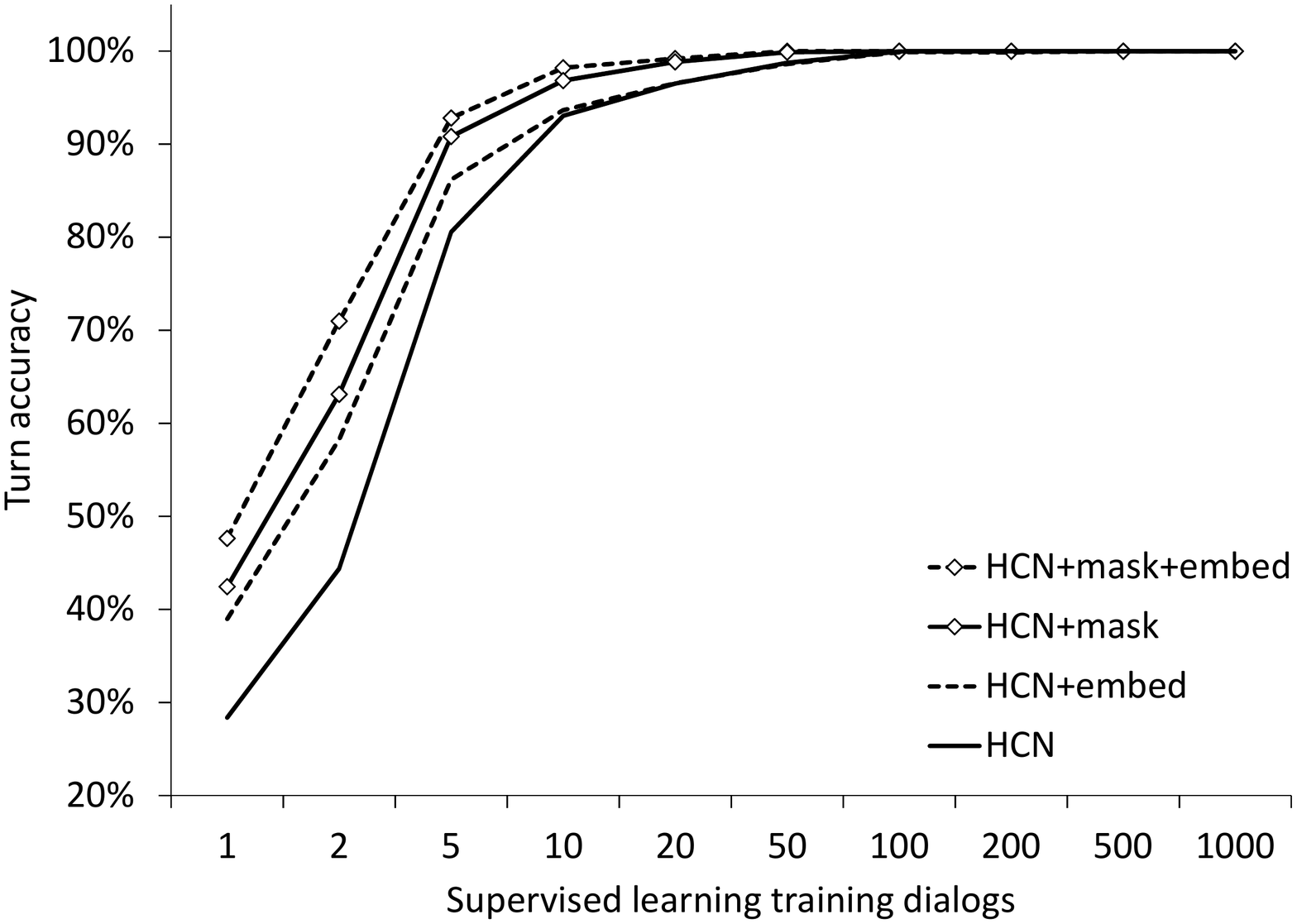}
        \caption{bAbI dialog Task5-OOV.}
    \end{subfigure}%
    ~ 
    \begin{subfigure}[t]{0.5\textwidth}
        \centering
        \includegraphics[trim = 20mm 20mm 20mm 20mm, scale=0.33]{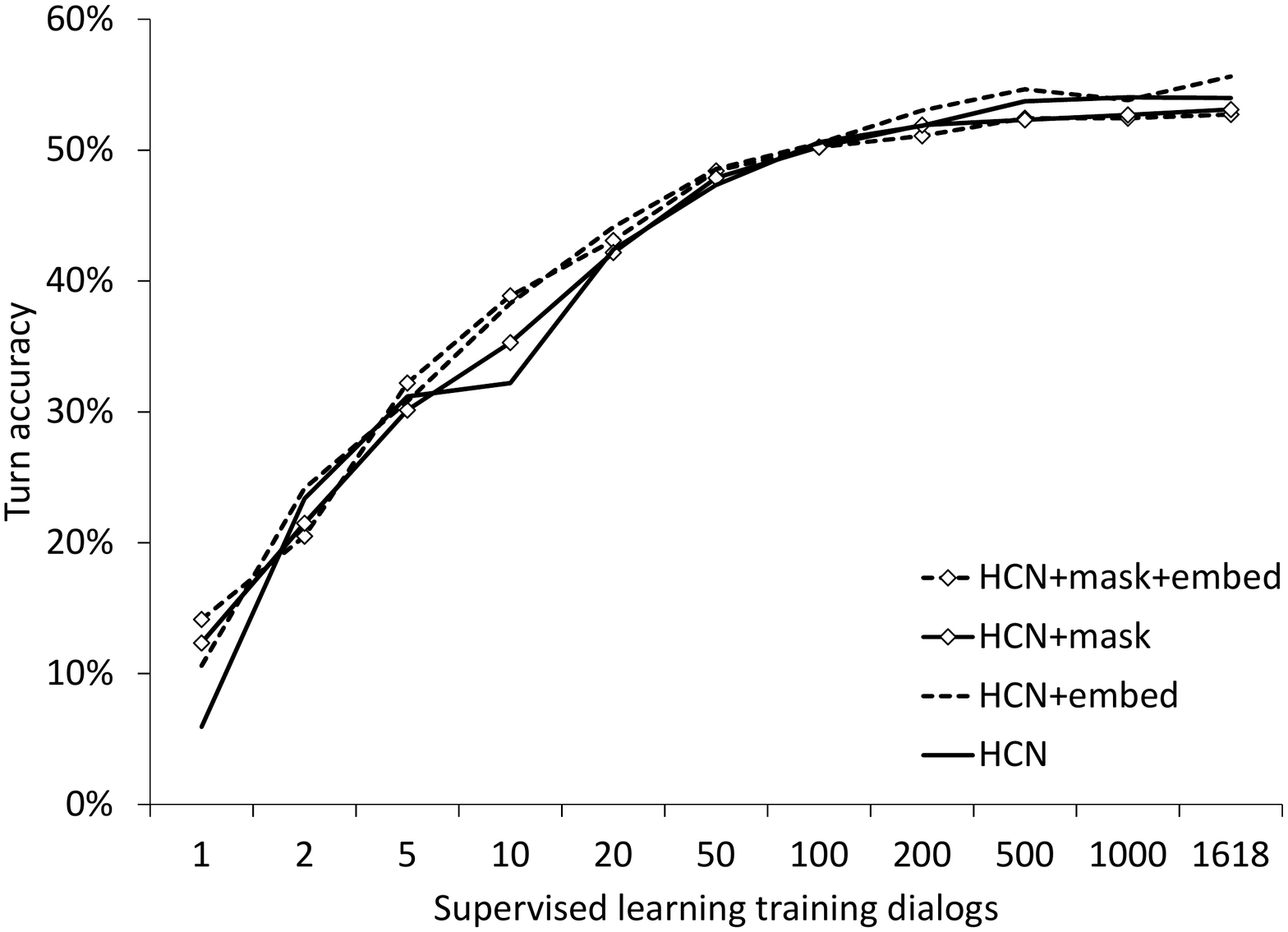}
        \caption{bAbI dialog Task6.}
    \end{subfigure}
    \caption{\label{fig:sl1}Training dialog count vs. turn accuracy for bAbI dialog Task5-OOV and Task6.  ``embed'' indicates whether utterance embeddings were included; ``mask'' indicates whether the action masking code was active.}
\end{figure*}

Results are shown in Table \ref{tab:sl1}.  Since Task5 is synthetic data generated using rules, it is possible to obtain perfect accuracy using rules (line 1).  The addition of domain knowledge greatly simplifies the learning task and enables HCNs to also attain perfect accuracy.  On Task6, rules alone fare poorly, whereas HCNs outperform past learned models.

We next examined learning curves, training with increasing numbers of dialogs.  To guard against bias in the ordering of the training set, we averaged over 5 runs, randomly permuting the order of the training dialogs in each run. Results are in Figure \ref{fig:sl1}.  In Task5, the action mask and utterance embeddings substantially reduce the number of training dialogs required (note the horizontal axis scale is logarithmic).  For Task6, the benefits of the utterance embeddings are less clear.  An error analysis showed that there are several systematic differences between the training and testing sets.  Indeed, DSTC2 intentionally used different dialog policies for the training and test sets, whereas our goal is to mimic the policy in the training set.  
%Also, in Task6, the input is the (error-free) transcription of the words spoken by the user, but the system's action is based on the (possibly erroneous) speech recognition output.  In other words, Task6 evaluates the ability to predict both dialog action \emph{and} speech recognition errors.

Nonetheless, these tasks are the best public benchmark we are aware of, and HCNs exceed performance of existing sequence-to-sequence models.  In addition, they match performance of past models using an order of magnitude less data (200 vs. 1618 dialogs), which is crucial in practical settings where collecting realistic dialogs for a new domain can be expensive.

\section{Supervised learning evaluation II}
\label{sec:sl2}

We now turn to comparing with purely hand-crafted approaches.  To do this, we obtained logs from our company's text-based customer support dialog system, which uses a sophisticated rule-based dialog manager.  Data from this system is attractive for evaluation because it is used by real customers -- not usability subjects -- and because its rule-based dialog manager was developed by customer support professionals at our company, and not the authors.  This data is not publicly available, but we are unaware of suitable human-computer dialog data in the public domain which uses rules.

Customers start using the dialog system by entering a brief description of their problem, such as ``I need to update my operating system''.  They are then routed to one of several hundred domains, where each domain attempts to resolve a particular problem.  In this study, we collected human-computer transcripts for the high-traffic domains ``reset password'' and ``cannot access account''.  

We labeled the dialog data as follows.  First, we enumerated unique system actions observed in the data.  Then, for each dialog, starting from the beginning, we examined each system action, and determined whether it was ``correct''.  Here, correct means that it was the most appropriate action among the set of existing system actions, given the history of that dialog.  If multiple actions were arguably appropriate, we broke ties in favor of the existing rule-based dialog manager.  Example dialogs are provided in the Appendix Sections \ref{sec:forgot_password_ed} and \ref{sec:account_access_ed}.

If a system action was labeled as correct, we left it as-is and continued to the next system action.  If the system action was not correct, we replaced it with the correct system action, and discarded the rest of the dialog, since we do not know how the user would have replied to this new system action.  The resulting dataset contained a mixture of complete and partial dialogs, containing only correct system actions.  We partitioned this set into training and test dialogs.  Basic statistics of the data are shown in Table \ref{tab:sl2}.

In this domain, no entities were relevant to the control flow, and there was no obvious mask logic since any question could follow any question.  Therefore, we wrote no domain-specific software for this instance of the HCN, and relied purely on the recurrent neural network to drive the conversation.  The architecture and training of the RNN was the same as in Section \ref{sec:sl1}, except that here we did not have enough data for a validation set, so we instead trained until we either achieved 100\% accuracy on the training set or reached 200 epochs.  

\begin{figure*}[t]
    \centering
    \begin{subfigure}[t]{0.5\textwidth}
        \centering
        \includegraphics[trim = 20mm 20mm 20mm 20mm, scale=0.33]{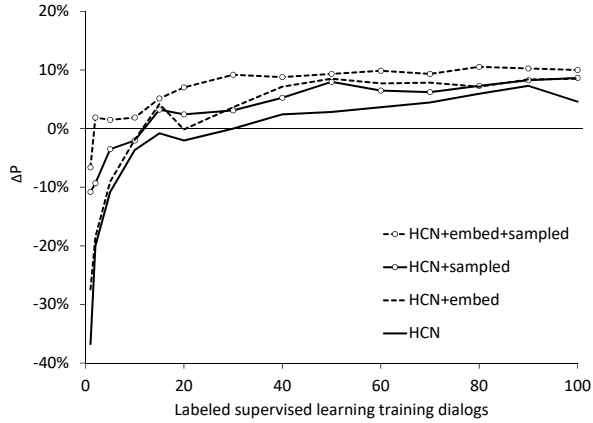}
        \caption{``Forgot password'' domain.}
    \end{subfigure}%
    ~ 
    \begin{subfigure}[t]{0.5\textwidth}
        \centering
        \includegraphics[trim = 20mm 20mm 20mm 20mm, scale=0.33]{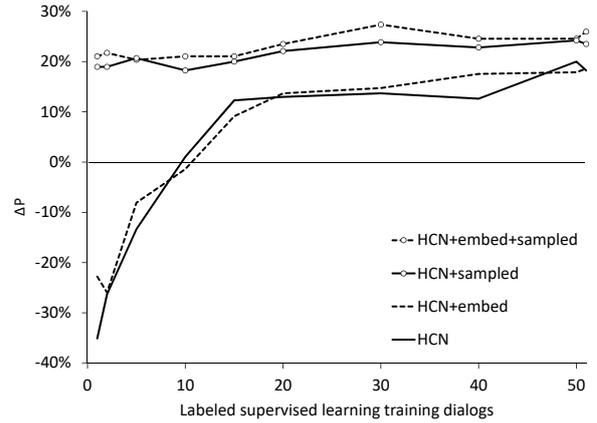}
        \caption{``Can't access account'' domain.}
    \end{subfigure}
    \caption{\label{fig:sl2}Training dialogs vs. $\Delta P$, where $\Delta P$ is the fraction of test dialogs where HCNs produced longer initial correct sequences of system actions than the rules, minus the fraction where rules produced longer initial correct sequences than the HCNs.  ``embed'' indicates whether utterance embeddings were included; ``sampled'' indicates whether dialogs sampled from the rule-based controller were included in the training set.}
\end{figure*}

To evaluate, we observe that conventional measures like average dialog accuracy unfairly penalize the system used to collect the dialogs -- in our case, the rule-based system.  If the system used for collection makes an error at turn $t$, the labeled dialog only includes the sub-dialog up to turn $t$, and the system being evaluated off-line is only evaluated on that sub-dialog.  In other words, in our case, reporting dialog accuracy would favor the HCN because it would be evaluated on fewer turns than the rule-based system.
% the rule-based system unfairly, since when the rule-based system makes an error in turn $t$, the HCN is only evaluated on the subsequence ending at turn $t$.  
We therefore use a comparative measure that examines which method produces longer continuous sequences of correct system actions, starting from the beginning of the dialog.  Specifically, we report $\Delta P = \frac{C(\text{HCN-win}) - C(\text{rule-win})}{C(\text{all})}$, where $C(\text{HCN-win})$ is the number of test dialogs where the rule-based approach output a wrong action before the HCN; $C(\text{rule-win})$ is the number of test dialogs where the HCN output a wrong action before the rule-based approach; and $C(\text{all})$ is the number of dialogs in the test set.  When $\Delta P > 0$, there are more dialogs in which HCNs produce longer continuous sequences of correct actions starting from the beginning of the dialog.  
We run all experiments 5 times, each time shuffling the order of the training set.  Results are in Figure \ref{fig:sl2}. HCNs exceed performance of the existing rule-based system after about 30 dialogs.
%, and in both domains the utterance embeddings accelerate learning, particularly in the ``forgot password'' domain.

%We evaluated as in Section \ref{sec:sl1}, and report results in terms of dialog accuracy -- i.e., a dialog is correct only if all turns are predicted correctly. 
%We evaluated our approach by training it on increasingly large subsets of the training set.  For each learned model, we evaluated it on the test set by determine whether it would produce the system actions in the dialog given the same sequence of user input.  If every system action in the labeled dialog matched the learned model output, that dialog was scored as ``error-free''.  If any predicted action differed from the labeled action, the dialog was scored as an ``error.''.  We repeated this process 10 times, each time shuffling the order of the training set, to remove any bias introduced by the order of the training set.  

\begin{table}[t]
\begin{center}
\begin{tabular}{c|cc}
\hline 
%  & \multicolumn{2}{|c}{\bf Domain}
  & Forgot & Account \\
  & password & Access \\
\hline
Av. sys. turns/dialog   & 2.2    & 2.2    \\
Max. sys. turns/dialog  &   5    &   9    \\
Av. words/user turn     & 7.7    & 5.4    \\
Unique sys. actions     &   7    &  16    \\
Train dialogs           & 422    & 56     \\
Test dialogs            & 148    & 60     \\
Test acc. (rules)       & 64.9\% & 42.1\% \\
\hline
\end{tabular}
\end{center}
\caption{\label{tab:sl2} Basic statistics of labeled customer support dialogs. Test accuracy refers to whole-dialog accuracy of the existing rule-based system.}
\end{table}

In these domains, we have a further source of knowledge: the rule-based dialog managers themselves can be used to generate example ``sunny-day'' dialogs, where the user provides purely expected inputs.  From each rule-based controller, synthetic dialogs were sampled to cover each expected user response at least once, and added to the set of labeled real dialogs.  This resulted in 75 dialogs for the ``Forgot password'' domain, and 325 for the ``Can't access account'' domain.  Training was repeated as described above.  Results are also included in Figure \ref{fig:sl2}, with the suffix ``sampled''.  In the ``Can't access account'' domain, the sampled dialogs yield a large improvement, probably because the flow chart for this domain is large, so the sampled dialogs increase coverage.  The gain in the ``forgot password'' domain is present but smaller. 

In summary, HCNs can out-perform production-grade rule-based systems with a reasonable number of labeled dialogs, and adding synthetic ``sunny-day'' dialogs improves performance further.  Moreover, unlike existing pipelined approaches to dialog management that rely on an explicit state tracker, this HCN used no explicit state tracker, highlighting an advantage of the model. 

\section{Reinforcement learning illustration}
\label{sec:rl}

In the previous sections, supervised learning (SL) was applied to train the LSTM to mimic dialogs provided by the system developer.  Once a system operates at scale, interacting with a large number of users, it is desirable for the system to continue to learn \emph{autonomously} using reinforcement learning (RL).  With RL, each turn receives a measurement of goodness called a \emph{reward}; the agent explores different sequences of actions in different situations, and makes adjustments so as to maximize the expected discounted sum of rewards, which is called the \emph{return}, denoted $G$.  

% \begin{figure}[t]
% \begin{center}
% \includegraphics[trim = 20mm 20mm 20mm 20mm, scale=0.34]{rl1.pdf}
% \caption{\label{fig:rl1}Dialog success rate vs. reinforcement learning training dialogs.  Each curve shows SL pre-training with a different number of labeled dialogs.}
% \end{center}
% \end{figure}

% \begin{figure}[t]
% \begin{center}
% \includegraphics[trim = 20mm 20mm 20mm 20mm, scale=0.34]{rl2.pdf}
% \caption{\label{fig:rl2}Dialog success rate vs. reinforcement learning training dialogs.  Curve marked 0 begins with a randomly initialized LSTM.  Curve marked 10 is pre-trained with 10 labeled dialogs.  Curve marked ``10 interleaved'' adds one SL training dialog before RL dialog 0, 100, 200, ... 900.}
% \end{center}
% \end{figure}

For optimization, we selected a \emph{policy gradient} approach \citep{williams1992simple}, which has been successfully applied to dialog systems \citep{jurvcivcek2011natural}, robotics \citep{kohl2004policy}, and the board game Go \citep{silver2016mastering}.  In policy gradient-based RL, a model $\pi$ is parameterized by $\mathbf{w}$ and outputs a distribution from which actions are sampled at each timestep.  At the end of a trajectory -- in our case, dialog -- the return $G$ for that trajectory is computed, and the gradients of the probabilities of the actions taken with respect to the model weights are computed.  The weights are then adjusted by taking a gradient step proportional to the return: 
\begin{equation}
\mathbf{w} \leftarrow \mathbf{w} + \alpha ( \sum_t \triangledown_{\mathbf{w}} \log \pi(a_t|\mathbf{h_t};\mathbf{w}) ) ( G - b ) \label{eq:policygrad}
\end{equation}  
where $\alpha$ is a learning rate; $a_t$ is the action taken at timestep $t$; $\mathbf{h_t}$ is the dialog history at time $t$; $G$ is the return of the dialog; $\triangledown_{\mathbf{x}} F$ denotes the Jacobian of $F$ with respect to $\mathbf{x}$; $b$ is a baseline described below; and $\pi(a|\mathbf{h};\mathbf{w})$ is the LSTM -- i.e., a stochastic policy which outputs a distribution over $a$ given a dialog history $\mathbf{h}$, parameterized by weights $\mathbf{w}$.  The baseline $b$ is an estimate of the average return of the current policy, estimated on the last 100 dialogs using weighted importance sampling.\footnote{The choice of baseline does not affect the long-term convergence of the algorithm (i.e., the bias), but can dramatically affect the speed of convergence (i.e., the variance) \citep{williams1992simple}.}  Intuitively, ``better'' dialogs receive a positive gradient step, making the actions selected more likely; and ``worse'' dialogs receive a negative gradient step, making the actions selected less likely.  

SL and RL correspond to different methods of updating weights, so both can be applied to the same network.  However, there is no guarantee that the optimal RL policy will agree with the SL training set; therefore, after each RL gradient step, we check whether the updated policy reconstructs the training set.  If not, we re-run SL gradient steps on the training set until the model reproduces the training set.  Note that this approach allows new training dialogs to be added at any time during RL optimization.  

%Past work has applied the so-called \emph{natural} gradient estimate \citep{peters2008natural} to dialog systems \citep{jurvcivcek2011natural}.  The natural gradient is a second-order gradient estimate which has often been shown to converge faster than the standard gradient.  However, computing the natural gradient requires inverting a matrix of model weights, which we found to be intractable for the large numbers of weights found in neural networks.  

%To the standard policy gradient update, we make three modifications.  First, the effect of the action mask is to clamp some action probabilities to zero, which causes the logarithm term in the policy gradient update to be undefined.  To solve this, we add a small constant to all action probabilities before applying the update.
%\footnote{The value of the constant is unimportant since the derivative in the update is unaffected by an additive constant.}  

We illustrate RL optimization on a simulated dialog task in the name dialing domain.  In this system, a contact's name may have synonyms (``Michael'' may also be called ``Mike''), and a contact may have more than one phone number, such as ``work'' or ``mobile'', which may in turn have synonyms like ``cell'' for ``mobile''.  This domain has a database of names and phone numbers taken from the Microsoft personnel directory, 5 entity types -- \verb|firstname|, \verb|nickname|, \verb|lastname|, \verb|phonenumber|, and \verb|phonetype| --  and 14 actions, including 2 API call actions.  Simple entity logic was coded, which retains the most recent copy of recognized entities.  A simple action mask suppresses impossible actions, such as placing a phonecall before a phone number has been retrieved from the database.  Example dialogs are provided in Appendix Section \ref{sec:name_dialing_ed}.

To perform optimization, we created a simulated user.  At the start of a dialog, the simulated user randomly selected a name and phone type, including names and phone types not covered by the dialog system.  When speaking, the simulated user can use the canonical name or a nickname; usually answers questions but can ignore the system; can provide additional information not requested; and can give up.  The simulated user was parameterized by around 10 probabilities, set by hand.

We defined the reward as being $1$ for successfully completing the task, and $0$ otherwise.  A discount of $0.95$ was used to incentivize the system to complete dialogs faster rather than slower, yielding return $0$ for failed dialogs, and $G = 0.95^{T-1}$ for successful dialogs, where $T$ is the number of system turns in the dialog. Finally, we created a set of 21 labeled dialogs, which will be used for supervised learning.

For the RNN in the HCN, we again used an LSTM with AdaDelta, this time with 32 hidden units.  RL policy updates are made after each dialog.  Since a simulated user was employed, we did not have real user utterances, and instead relied on context features, omitting bag-of-words and utterance embedding features.

We first evaluate RL by randomly initializing an LSTM, and begin RL optimization.  After 10 RL updates, we freeze the policy, and run 500 dialogs with the user simulation to measure task completion.  We repeat all of this for 100 runs, and report average performance.  In addition, we also report results by initializing the LSTM using supervised learning on the training set, consisting of 1, 2, 5, or 10 dialogs sampled randomly from the training set, then running RL as described above.

\begin{figure}[t]
\begin{center}
\includegraphics[trim = 20mm 20mm 20mm 20mm, scale=0.33]{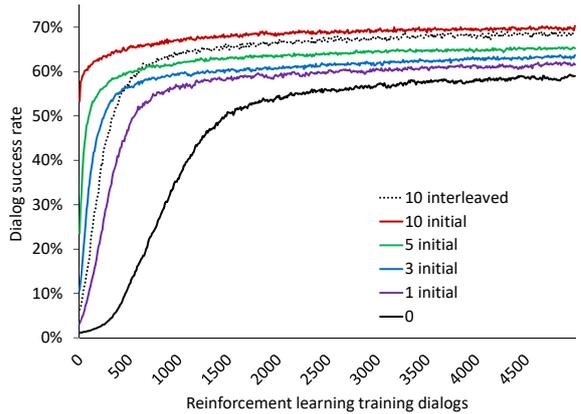}
\caption{\label{fig:rl2a}Dialog success rate vs. reinforcement learning training dialogs.  Curve marked ``0'' begins with a randomly initialized LSTM.  Curves marked ``$N$ initial'' are pre-trained with $N$ labeled dialogs.  Curve marked ``10, interleaved'' adds one SL training dialog before RL dialog 0, 100, 200, ... 900.}
\end{center}
\end{figure}

Results are in Figure \ref{fig:rl2a}.  Although RL alone can find a good policy, pre-training with just a handful of labeled dialogs improves learning speed dramatically.  Additional experiments, not shown for space, found that ablating the action mask slowed training, agreeing with \citet{williams2008icslp1}.

Finally, we conduct a further experiment where we sample 10 training dialogs, then add one to the training set just before RL dialog 0, 100, 200, ... , 900.  Results are shown in Figure \ref{fig:rl2a}.  This shows that SL dialogs can be introduced as RL is in progress -- i.e., that it is possible to interleave RL and SL.  This is an attractive property for practical systems: if a dialog error is spotted by a developer while RL is in progress, it is natural to add a training dialog to the training set. 

\section{Conclusion}
\label{sec:concl}

This paper has introduced Hybrid Code Networks for end-to-end learning of task-oriented dialog systems.  HCNs support a separation of concerns where procedural knowledge and constraints can be expressed in software, and the control flow is learned.  Compared to existing end-to-end approaches, HCNs afford more developer control and require less training data, at the expense of a small amount of developer effort.  

Results in this paper have explored three different dialog domains.  On a public benchmark in the restaurants domain, HCNs exceeded performance of purely learned models.  Results in two troubleshooting domains exceeded performance of a commercially deployed rule-based system.  Finally, in a name-dialing domain, results from dialog simulation show that HCNs can also be optimized with a mixture of reinforcement and supervised learning.

In future work, we plan to extend HCNs by incorporating lines of existing work, such as integrating the entity extraction step into the neural network \cite{dhingra2017}, adding richer utterance embeddings \cite{socher2013}, and supporting text generation \cite{sordoni2015}.  We will also explore using HCNs with automatic speech recognition (ASR) input, for example by forming features from n-grams of the ASR n-best results \cite{Henderson2014b}.  Of course, we also plan to deploy the model in a live dialog system.  More broadly, HCNs are a general model for stateful control, and we would be interested to explore applications beyond dialog systems -- for example, in NLP medical settings or human-robot NL interaction tasks, providing domain constraints are important for safety; and in resource-poor settings, providing domain knowledge can amplify limited data.

\bibliography{master}
\bibliographystyle{acl_natbib}

\appendix

\section{Supplemental Material}
\label{sec:supplemental}

\subsection{Model implementation details}
\label{sec:impl_details}

The RNN was specified using Keras version 0.3.3, with back-end computation in Theano version 0.8.0.dev0 \citep{2016arXiv160502688short,chollet2015keras}.  The Keras model specification is given below.  The input variable \verb|obs| includes all features from Figure \ref{fig:diagram} step 6 except for the previous action (step 18) and the action mask (step 6, top-most vector).

\begin{Verbatim}[fontsize=\small]
# Given: 
# obs_size, action_size, nb_hidden

g = Graph()
g.add_input(
  name='obs',
  input_shape=(None, obs_size)
)
g.add_input(
  name='prev_action',
  input_shape=(None, action_size)
)
g.add_input(
  name='avail_actions',
  input_shape=(None, action_size)
)
g.add_node(
  LSTM(
    n_hidden,
    return_sequences=True,
    activation='tanh',
  ),
  name='h1',
  inputs=[
    'obs',
    'prev_action',
    'avail_actions'
  ]
)
g.add_node(
  TimeDistributedDense(
    action_size,
    activation='softmax',
  ),
  name='h2',
  input='h1'
)
g.add_node(
  Activation(
    activation=normalize,
  ),
  name='action',
  inputs=['h2','avail_actions'],
  merge_mode='mul',
  create_output=True
)
g.compile(
  optimizer=Adadelta(clipnorm=1.),
  sample_weight_modes={
    'action': 'temporal'
  },
  loss={
    'action':'categorical_crossentropy'
  }
)
\end{Verbatim}

Model sizes are given in Table \ref{tab:model_details}. Example dialogs are given below for each of the 5 dialog systems.  For space and readability, the entity tags that appear in the user and system sides of the dialogs have been removed -- for example, {\sffamily {\emph{Call} \verb|<name>|\emph{Joan}\verb|</name>|}} is shown as {\sffamily \emph{Call Joan}}.

\begin{table*}[t]
\begin{center}
\begin{tabular}{l|ccccc}
\hline 
                               & \bf bAbI Task5 & \bf bAbI Task6 & \bf Reset password & \bf Account access & \bf Dialer \\
\hline 
\hspace{5mm} Utterance embed.  & 300            & 300            & 300                & 300                & 0           \\
\hspace{5mm} Bag of words      & 85             & 523            & 1124               & 278                & 0           \\ 
\hspace{5mm} Context features  & 4              & 14             & 0                  & 0                  & 17          \\ 
\verb|obs_size|                & 389            & 837            & 1424               & 578                & 17            \\
\hline
\verb|action_size|             & 16             & 58             & 7                  & 16                 & 14          \\ 
\hline
\verb|nb_hidden|               & 128            & 128            & 128                & 128                & 32          \\ 
\hline
\end{tabular}
\end{center}
\caption{\label{tab:model_details} Dimensions of the 5 HCNs in this paper.}
\end{table*}

\subsection{bAbI Task5 example dialog}
\label{sec:task5_ed}

\begin{sffamily}
\begin{framed}
\begin{small}
\noindent \emph{good morning}

\noindent hello what can i help you with today

\noindent \emph{i'd like to book a table with italian food}

\noindent i'm on it

\noindent \emph{[silence]}

\noindent where should it be

\noindent \emph{in paris}

\noindent how many people would be in your party

\noindent \emph{for six people please}

\noindent which price range are looking for

\noindent \emph{in a cheap price range please}

\noindent ok let me look into some options for you

\noindent \emph{[silence]}

\noindent \verb|api_call italian paris six cheap|

\noindent \verb|[database results returned]|

\noindent \emph{[silence]}

\noindent what do you think of this option: resto\_madrid\_cheap\_spanish\_8stars

\noindent \emph{no this does not work for me}

\noindent sure let me find an other option for you

\noindent \emph{[silence]}

\noindent what do you think of this option: resto\_madrid\_cheap\_spanish\_6stars

\noindent \emph{do you have something else}

\noindent sure let me find an other option for you

\noindent \emph{[silence]}

\noindent what do you think of this option: resto\_madrid\_cheap\_spanish\_1stars

\noindent \emph{it's perfect}

\noindent great let me do the reservation

\noindent \emph{may i have the phone number of the restaurant}

\noindent here it is resto\_madrid\_cheap\_spanish\_1stars\_phone

\noindent \emph{thanks}

\noindent is there anything i can help you with

\noindent \emph{no thank you}

\noindent you're welcome

\end{small}
\end{framed}
\end{sffamily}

\subsection{bAbI Task6 example dialog}
\label{sec:task6_ed}

In the system turn marked [*], the system provided venue information without first making an API call.  

\begin{sffamily}
\begin{framed}
\begin{small}

\noindent Hello , welcome to the Cambridge restaurant system . You can ask for restaurants by area , price range or food type . How may I help you ?

\noindent \emph{im looking for canapes food in any price range}

\noindent I am sorry but there is no canapes restaurant that matches your request [*]

\noindent \emph{how about european}

\noindent \verb|api_call european R_location R_price|

\noindent \verb|[database results returned]|

\noindent \emph{[silence]}

\noindent michaelhouse\_cafe is a great restaurant serving european food and it is in the expensive price range

\noindent \emph{food an}

\noindent What part of town do you have in mind?

\noindent \emph{any any part of town}

\noindent michaelhouse\_cafe serves european food .

\noindent \emph{may i have the address and phone number please}

\noindent The phone number of michaelhouse\_cafe is michaelhouse\_cafe\_phone

\noindent \emph{thank you good bye}

\noindent you are welcome

\end{small}
\end{framed}
\end{sffamily}

\subsection{Task5 and Task6 context features}
\label{sec:context_features}

Context features used in Section \ref{sec:sl1} are in Table \ref{tab:context_features}.

\begin{table*}[t]
\begin{center}
\begin{tabular}{l|cc}
\hline 
Feature                                                     & \bf bAbI Task5 & \bf bAbI Task6 \\
\hline 
Presence of each entity in dialog state                     & 4              & 3              \\
Presence of each entity in current utterance                & --             & 3              \\
Whether DB has been queried yet                             & --             & 1              \\
Whether DB results are empty                                & --             & 1              \\
Whether DB results are non-empty                            & --             & 1              \\
Whether any DB results have been presented                  & --             & 1              \\
Whether all DB results have been presented                  & --             & 1              \\
Whether any DB results are available to present             & --             & 1              \\
Whether curent query yields no results in training set             & --             & 1              \\
Whether cuisine in current query yields no results in training set & --             & 1              \\
\hline
Total context features                                      & 4              & 14             \\
\hline 
\end{tabular}
\end{center}
\caption{\label{tab:context_features} Binary context features used to convey entity and database state in Section \ref{sec:sl1}.}
\end{table*}

\subsection{Forgot password example dialog}
\label{sec:forgot_password_ed}

Some system actions have been shortened for space.

\begin{sffamily}
\begin{framed}
\begin{small}

\noindent \emph{How do I recover my password}

\noindent Okay, you don't need to remember your password, we can reset it. Would you like to try that?

\noindent \emph{Yes}

\noindent To reset your password, please see this URL ... Was that helpful?

\noindent \emph{I don't have a flashdrive}

\noindent Let's connect you to a person who can help you

\end{small}
\end{framed}
\end{sffamily}

\subsection{Account access example dialog}
\label{sec:account_access_ed}

Some system actions have been shortened for space.

\begin{sffamily}
\begin{framed}
\begin{small}

\noindent \emph{cannot sign in to my email}

\noindent Please check caps lock is off, and that your email is spelled correctly. Are you able to access your account now?

\noindent \emph{no}

\noindent If you still can't sign in, please go the the following page and check that the email server is running ok... Is everything up and running?

\noindent \emph{yes}

\noindent You need to reset your password - here's how ... Did this solve your problem?

\noindent \emph{still cannot get in}

\noindent Did you receive the security code?

\noindent \emph{yes}

\noindent Check you entered the right numbers. Did this solve your problem?

\noindent \emph{no}

\noindent Have you seen a message that your profile is temporarily blocked?

\noindent \emph{just your message of update on sept 15}

\noindent Did you turn on two-step verification recently?

\noindent \emph{whats that?}

\noindent Sorry, I can't help. Rephrase your question, or click below to talk to a person.

\end{small}
\end{framed}
\end{sffamily}

\subsection{Name dialing example dialogs}
\label{sec:name_dialing_ed}

\begin{sffamily}
\begin{framed}
\begin{small}
\noindent How can I help you?

\noindent \emph{Call Joe}

\noindent There's more than one person named Joe. Can you say their full name?

\noindent \emph{Call Joe Adamson}

\noindent \texttt{SavePhonetypeavail()}

\noindent Calling Joe Adamson, work

\noindent \texttt{PlaceCall()}

\end{small}
\end{framed}

\begin{framed}
\begin{small}
\noindent How can I help you?

\noindent \emph{Call Sally on her home}

\noindent Sorry, I don't have a home number for Sally Smith.  I only have a work phone.  Do you want to call that number?

\noindent \emph{no}

\noindent Oh, sorry about that.  Goodbye.

\end{small}
\end{framed}
\end{sffamily}

\end{document}